\documentclass[letterpaper, 10 pt, conference]{ieeeconf}  

\IEEEoverridecommandlockouts                              

\usepackage{graphicx}
\graphicspath{{figures/}}
\usepackage{multirow}
\usepackage{url}
\usepackage{amsmath} 
\usepackage{amsfonts}
\usepackage{amssymb}
\usepackage{bm}
\usepackage{booktabs}
\usepackage{siunitx}
\let\labelindent\relax
\usepackage{enumitem}
\usepackage[caption=false,font=footnotesize]{subfig}
\usepackage{svg}
\usepackage{cite}

\title{\LARGE \bf
Dynamics Modeling of a Multi-UAV Slung Load System Using a Discrete-Link Cable Approach
}

\author{Harvey Merton$^{1}$, and Ian W. Hunter$^{1}$
\thanks{$^{1}$The authors are with the BioInstrumentation Laboratory, Massachusetts
Institute of Technology, Cambridge, MA 02139, United States of America
(email: hmer101@mit.edu; ihunter@mit.edu)}
}

\begin{document}

\maketitle
\thispagestyle{empty}
\pagestyle{empty}

\begin{abstract}
A common assumption to simplify the problem of controlling a multi-UAV slung load system (MUSLS) is that the flexible cables can be modeled as massless rigid rods. In this work, we propose an alternative Euler-Newton derived dynamical model which uses a series of rigid links to model the flexible cables. The model is specifically designed to allow efficient simulation using Featherstone's articulated body algorithm. We perform real-world validation of this model on gentle, aggressive, and tension-engagement maneuvers and run a parameter sweep to determine the number of links, joint damping, and joint friction to achieve the greatest model fidelity. The model closely matches real-world flight data with mean load translation errors below 132 mm (5.5\% of the cable length) and orientation errors below 11.4 degrees. We make the real-world flight data publicly available for the development of future cable models.

\end{abstract}



\section*{Supplementary Material}\label{sect:supplementary_material}
\textbf{Video}: \verb|https://youtu.be/VATTPYYDwYY|

\textbf{Code}:\verb| https://github.com/hmer101/|
                \verb|musls_cable_modeling/|

\section{INTRODUCTION}\label{sect:intro}

\begin{figure}[htbp]
    \centerline{\includegraphics[width=1\columnwidth]{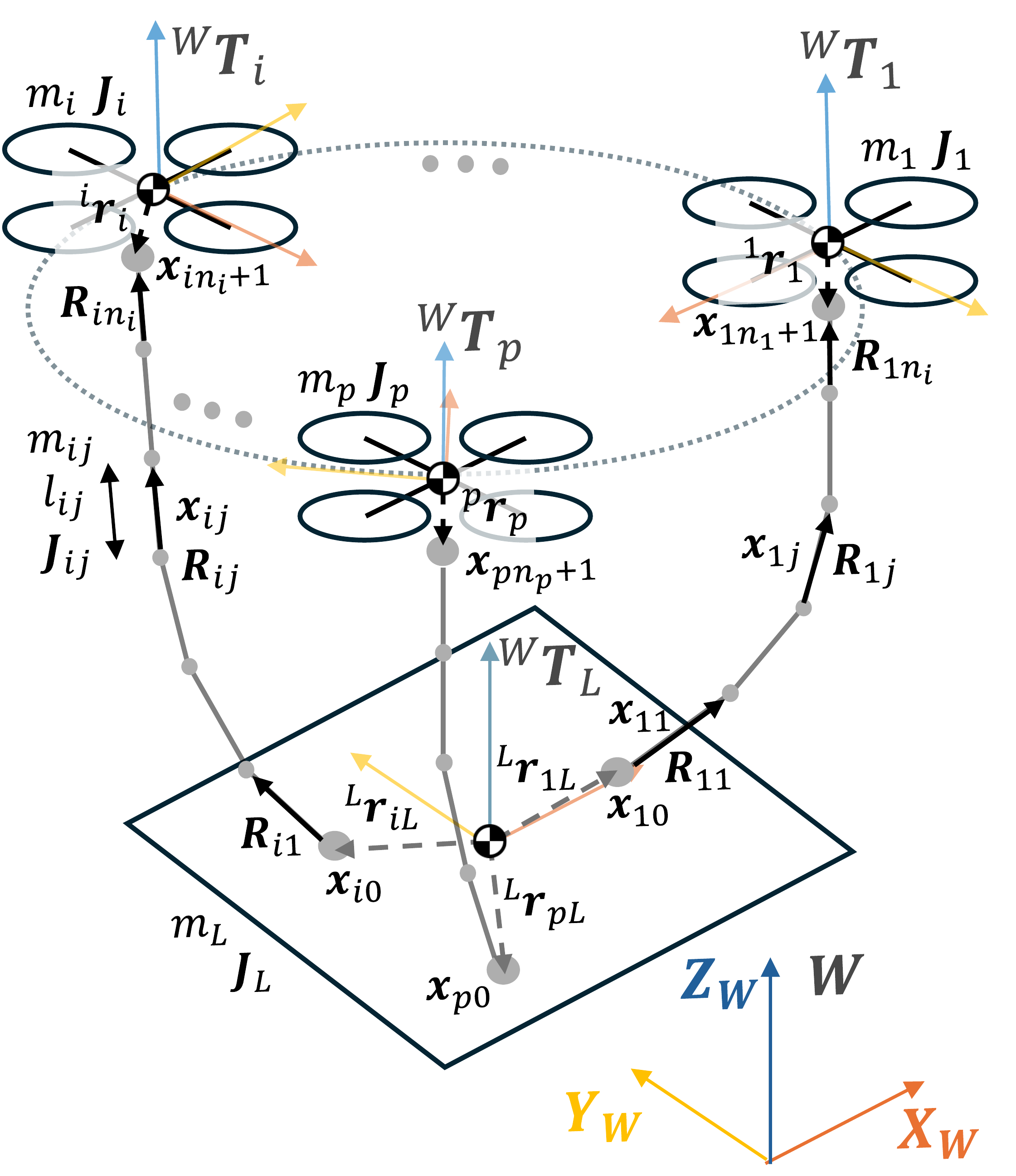}}
    \caption{A multi-UAV slung load system using a discrete-link cable model. This example features $p=3$ drones, where each cable attaching the $i^{\text{th}}$ drone to the load has $n_i$ links. The symbols used are defined in Section~\ref{sect:dynamics}.}
    \label{fig:system_symbols}
\end{figure}

A common strategy for unmanned aerial vehicle (UAV) payload transport is to suspend the payload from a flexible cable, yielding a slung-load configuration \cite{marques_tethered_2023}. When multiple UAVs cooperate to carry a single load (Fig.~\ref{fig:system_symbols}), they can transport heavier payloads and, in principle, regulate the load’s full six-degree-of-freedom pose \cite{manubens_motion_2013}.

As applications mature, there is a growing need to simulate multi-UAV slung-load systems (MUSLS) within off-the-shelf toolchains. In these settings, the dynamics model must be plug-and-play with standard rigid-body simulators, operate near real time, and remain stable across regimes that include loss and re-engagement of cable tension.

Accurately modeling general flexible cables is difficult because they exhibit nonlinear, high-dimensional dynamics and undergo abrupt transitions when tension is lost or re-established. Approaches to cable modeling can be categorized as discrete or continuous \cite{lv_review_2020}. \emph{Discrete} approaches include mass–spring systems \cite{lv_physically_2017}, multibody chains of rigid links \cite{servin_rigid_2008}, and finite-element methods \cite{wang_efficient_2013}; \emph{continuous} formulations include elastic-rod and spline-based dynamics \cite{lang_multi-body_2011, echegoyen_linked_2010}.

Within the MUSLS literature specifically, prior work can be grouped into three subsets of the discrete approach. \emph{Massless rigid links.} The most common approach assumes each cable is a massless, always-taut rod forming a straight line between attachment points \cite{sreenath_dynamics_2013, fink_planning_2011, michael_cooperative_2011, lee_geometric_2013, wu_geometric_2014}. This simplifies analysis and controller design but reduces simulation fidelity, particularly when cables go slack. \emph{Hybrid tension models.} To handle slack-cable transitions, some works augment rigid-link models with hybrid dynamics that explicitly switch between taut and slack regimes \cite{sreenath_dynamics_2013, li_rotortm_2024}. While more physically accurate during tension re-engagement, these models still rely on custom numerical simulators and are not easily integrated with general-purpose robotics toolchains. \emph{Discrete rigid links.} Discrete-link formulations have been applied to MUSLS primarily for differential-flatness-based planning and LQR control \cite{kotaru_differential-flatness_2018, kotaru_multiple_2020} and for stabilizing geometric control \cite{goodarzi_stabilization_2015}. These works emphasize control-theoretic insights rather than validating cable-model fidelity against real-world MUSLS data, and they offer no insight into how joint properties affect predictive performance. Moreover, all prior discrete-link MUSLS works concentrate link masses at joint connections rather than distributing inertia along each link. While this simplifies the state representation, it increases the number of elements required for convergence relative to rod-inertia models \cite{quisenberry_discrete_2006} and prevents treating links as standard rigid bodies. This is a key obstacle to integration with general-purpose simulators, most of which target tree-structured rigid bodies solved with Featherstone's articulated-body algorithm (ABA) \cite{featherstone_rigid_2008}.

\subsection{Organization and contributions}
This paper focuses on the construction and performance quantification of a multi-discrete-link cable dynamics model in a MUSLS context. The model is specifically designed to be easily integrated with a range of rigid-body robotics simulators. The key contributions are:

\begin{itemize}
    \item a dynamical model of a MUSLS with a discrete-link cable model which includes the effect of link inertias and tunable joint properties, 
    \item a systematic selection of key parameters to balance computational efficiency and predictive performance of a multi-link cable model, 
    \item a quantification of the performance of a multi-link cable model over a variety of trajectories (including tension engagement), and
    \item real-world flight data made publicly available \cite{harvey_merton_code_2025} to allow the development of future cable models.
\end{itemize}

The remainder of the paper is organized as follows. Section \ref{sect:dynamics} introduces the MUSLS dynamical model, with emphasis on the multi-link cable formulation and how it differs from existing approaches. Section \ref{sect:method} describes the approach to testing the model with a position control based scheme. Section \ref{sect:results} discusses model tuning and the performance of a multi-rigid-link cable model for MUSLS simulation. Section \ref{sect:conclusions} provides concluding remarks.


\section{DYNAMICS MODELING}\label{sect:dynamics}
We derive the equations of motion for the MUSLS in Fig.~\ref{fig:system_symbols} using the Euler–Newton approach. Unlike many prior formulations, the system is modeled as a tree of rigid bodies, making it solvable with the ABA. The section opens with notation and conventions, proceeds with the derivation of the equations of motion, and concludes with remarks comparing this formulation to existing MUSLS models.

\emph{Notation:} In this paper, scalars are written in Roman (e.g., $x$), vectors in bold lowercase (e.g., $\bm{x}$), and matrices in bold uppercase (e.g., $\bm{X}$). The special orthogonal ($SO(3)$) and special Euclidean ($SE(3)$) Lie groups are used to represent  orientations and rigid-body frame transformations respectively. They are defined as
\[
SO(3) := \{ \bm{R} \in \mathbb{R}^{3\times 3} \;\mid\; \bm{R}^\top \bm{R} = \bm{I},\; \det(\bm{R}) = +1 \},
\]
\[
SE(3) := \left\{ \;
\bm{T} =
\begin{bmatrix}
\bm{R} & \bm{t} \\
\bm{0} & 1
\end{bmatrix}
\;\middle|\; \bm{R} \in SO(3),\; \bm{t} \in \mathbb{R}^3
\;\right\}.
\]
The \emph{hat map} $\hat\cdot : \mathbb{R}^3 \to \mathfrak{so}(3)$ 
provides the isomorphism between $\mathbb{R}^3$ and the Lie algebra 
$\mathfrak{so}(3)$, defined by
\[
\hat{\bm{x}}\, \bm{y} = \bm{x} \times \bm{y}, 
\quad \forall\, \bm{x}, \bm{y} \in \mathbb{R}^3.
\] The notation $^a\bm{T}_b \in SE(3)$ represents the pose (position and orientation) of b in frame a. Vectors are taken relative to the origin of the coordinate frame used (e.g. frame a's origin in $^a\bm{r}_b$). If no frame is specified (e.g. $\bm{x}_{i0}$), the inertial coordinate frame ($W$ in Fig. \ref{fig:system_symbols}) is used. 

\emph{Definitions:} Consider a rigid body payload of mass $m_L$ and inertia $\bm{J}_L$ connected to $p$ drones through flexible cables, as shown in Fig. \ref{fig:system_symbols}. The position $\bm{x}_{L} \in \mathbb{R}^3$ (of the center of mass) and orientation $\bm{R}_L \in SO(3)$ of the load together define its pose in the inertial frame $^W\bm{T}_L \in SE(3)$. Each drone $i\in \{1,...,p\}$ can be described similarly with mass $m_i$, inertia $\bm{J}_i$, position $\bm{x}_i$, orientation $\bm{R}_i$, and pose $^W\bm{T}_i$. The drones modeled here are planar quadcopters capable of producing thrust $f_i \in \mathbb{R}$ perpendicular to the plane of the rotors, and moment $^i\bm{\tau}_i \in \mathbb{R}^3$ measured in the drone's body-fixed frame $^W\bm{T}_i$. The angular velocity of the load ($^L\bm{\omega}_L \in \mathbb{R}^3$) and drones ($^i\bm{\omega}_i \in \mathbb{R}^3$) respectively are defined in their body-fixed frames. 

\begin{figure}[htbp]
    \centerline{\includegraphics[width=1\columnwidth]{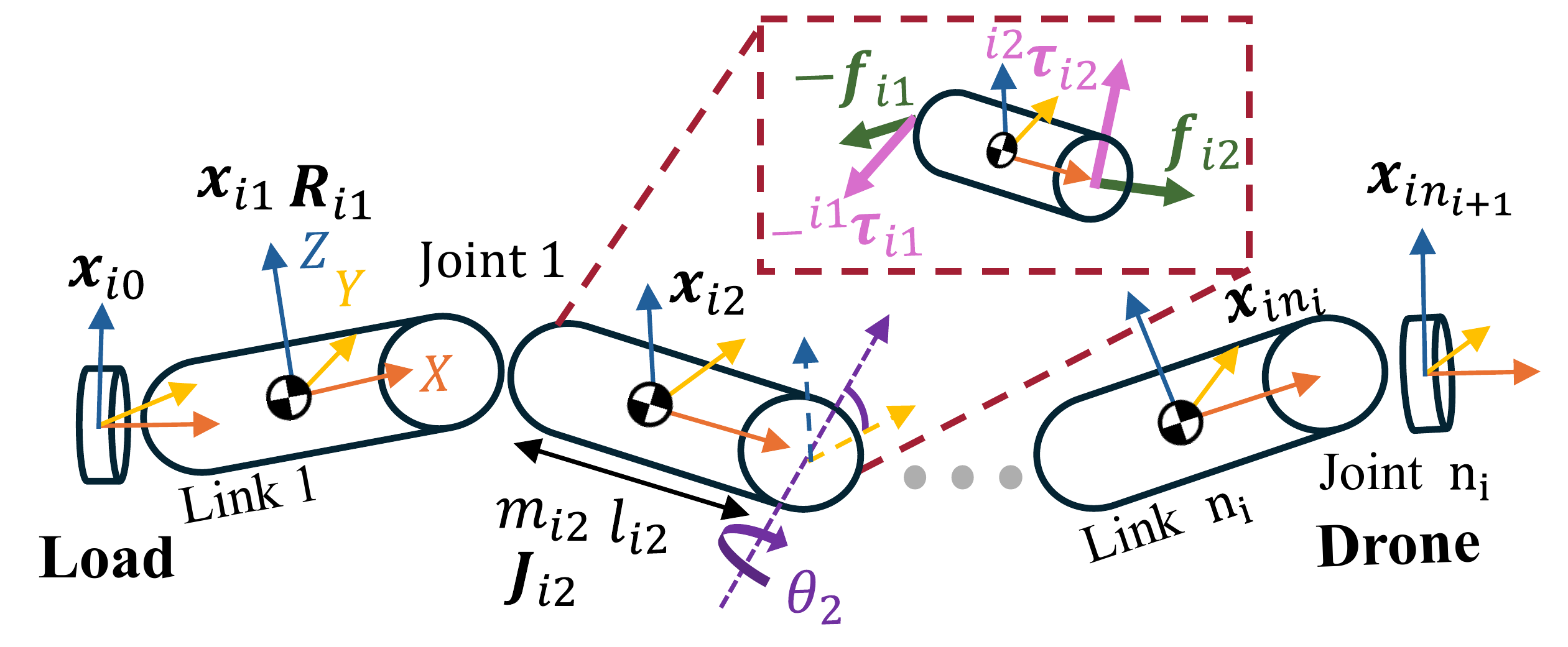}}
    \caption{A model for cable $i$ comprised of $n_i$ cylindrical links each with mass $m_{ij}$, length $l_{ij}$ and inertia $\bm{J}_{ij}$. The labeled joint axes are fixed at the end of each link, while the purple axis represents a generic axis around which link $j+1$ can rotate relative to link $j$ using a universal joint.} 
    \label{fig:cable_links}
\end{figure}

Each cable $i$ is modeled by a series of $n_i$ rigid cylinders of length $l_{ij}$ numbered $j=1,...,n_i$ (see Fig. \ref{fig:cable_links}). Similar to the load and drone rigid bodies, each link has mass $m_{ij}$ (at the center of the cylinder), inertia $\bm{J}_{ij}$, position $\bm{x}_{ij}$, orientation $\bm{R}_{ij}$, angular velocity $^{ij}\bm{\omega}_{ij}$, and pose $^W\bm{T}_{ij}$. The internal moments $^{ij}\bm{\tau}_{ij} \in \mathbb{R}^3$ between the joint links are defined in link $ij$'s frame, while the internal forces $\bm{f}_{ij} \in \mathbb{R}^3$ are defined in the inertial frame for convenience. 

\emph{Kinematic constraints:} The connection point of link $1$ in cable $i$ to the load is defined relative to the load's center of mass by $^L\bm{r}_{iL} \in \mathbb{R}^3$. Similarly, the connection point of link $n_{i}$ to the drone is defined relative to the drone's center of mass by $^i\bm{r}_{i} \in \mathbb{R}^3$. Links $j=0$ and $j=n_{i+1}$ represent the cable's connection points on the load and drone $i$ respectively. The kinematic relations between the load, drones, and cable links are:

\begin{align}
    &\bm{x}_{i0} = \bm{x}_{L} + \bm{R}_L{^L\bm{r}_{iL}}, \label{eq:cable_load_end}
    \\
    &\bm{x}_{ij} = \bm{x}_{i(j-1)} + \bm{R}_{i(j-1)}   
    \begin{bmatrix}
    l_{i(j-1)}/2\\
    0\\
    0
    \end{bmatrix} + \bm{R}_{ij}   
    \begin{bmatrix}
    l_{ij}/2\\
    0\\
    0
    \end{bmatrix},
    \label{eq:cable_mid_link}
    \\
    &\bm{x}_{i} = \bm{x}_{in_{i+1}}  - \bm{R}_i{^i\bm{r}_{i}}.\label{eq:cable_drone_end}
\end{align}

\emph{Dynamics:} The following assumptions are made to develop a dynamical model for the MUSLS described above:

\begin{enumerate}[label=(A\arabic*)]
    \item \label{ass:aero} aerodynamic forces acting on each cable are negligible,
    \item \label{ass:inelastic} each cable is inextensible (no axial stretch) and has negligible bending stiffness, and
    \item \label{ass:twisting_moment} the moment due to the twist of a cable around its own axis is negligible. 
\end{enumerate} Using the Euler–Newton method with these assumptions, we derive the following system of equations:

\begin{align}
    &m_L(\ddot{\bm{x}}_L + g {\bm{e}}_3) =  \sum_{i=1}^{p} \bm{f}_{i0}, \label{eq:eom_load_lin}
    \\
    &\bm{J}_L {^L\dot{\bm{\omega}}_L} + {^L\hat{\bm{\omega}}_L} \bm{J}_L {^L{\bm{\omega}}_L} \nonumber\\
    &= \sum_{i=1}^{p}\left( {^L\bm{r}_{iL}} \times \bm{R}^{\intercal}_L\bm{f}_{i0} + {^{i0}\bm{\tau}_{i0}}\right),\label{eq:eom_load_ang}
    \\
    &m_i(\ddot{\bm{x}}_{i} + g {\bm{e}}_3) = f_i \bm{R}_i {\bm{e}}_3  - \bm{f}_{in_i}, \label{eq:eom_drones_lin}
    \\
    &\bm{J}_i  {^i\dot{\bm{\omega}}_i} + {^i\hat{\bm{\omega}}_i} \bm{J}_i {^i\bm{\omega}}_i = {^i\bm{\tau}_i} - {^i\bm{r}_{i}} \times \bm{R}^{\intercal}_i\bm{f}_{in_i} - {^{in_i}\bm{\tau}_{in_i}}, \label{eq:eom_drones_ang}
    \\
    &m_{ij}(\ddot{\bm{x}}_{ij}+g\bm{e}_3) = \bm{f}_{ij} - \bm{f}_{i(j-1)},\label{eq:eom_cable_lin}
    \\
    &\bm{J}_{ij} {^{ij}\dot{\bm{\omega}}_{ij}} + {^{ij}\hat{\bm{\omega}}_{ij}} 
    \bm{J}_{ij} {^{ij}{\bm{\omega}}_{ij}} \nonumber \\
    &= 
    \begin{bmatrix}
    \frac{l_{ij}}{2}\\
    0\\
    0
    \end{bmatrix}
    \times 
    \bm{R}^{\intercal}_{ij}\bm{f}_{ij}
    +
    \begin{bmatrix}
    \frac{l_{ij}}{2}\\
    0\\
    0
    \end{bmatrix}
    \times 
    \bm{R}^{\intercal}_{i(j-1)}\bm{f}_{i(j-1)} \nonumber \\
    &\quad+{^{ij}\bm{\tau}_{ij}}
    - {^{i(j-1)}\bm{\tau}_{i(j-1)}}, \label{eq:eom_cable_ang}
\end{align} where $\bm{e}_1, \bm{e}_2, \bm{e}_3 \in \mathbb{R}^3$ are unit vectors along the $X$, $Y$ and $Z$ axes in the inertial frame. Here, (\ref{eq:eom_load_lin}) and (\ref{eq:eom_load_ang}) describe the translational and rotational dynamics of the load respectively. Similarly, (\ref{eq:eom_drones_lin}) and (\ref{eq:eom_drones_ang}) describe the dynamics of drone $i$, while the $j^{th}$ link of each cable $i$ is described by (\ref{eq:eom_cable_lin}) and (\ref{eq:eom_cable_ang}). The components of $^{ij}\bm{\tau}_{ij} \in \mathbb{R}^3$ define the moment generated around each joint $j$'s $X$, $Y$ and $Z$ axis (see Fig. \ref{fig:cable_links} for coordinate frame definition) respectively. The moment around each joint's $X$ axis is always \SI{0}{\newton\metre} due to \ref{ass:twisting_moment}. 

\emph{Joint model:} The joint moments $^{ij}\bm{\tau}_{ij}$ are generated by a combination of joint stiffness ($\bm{K} \in \mathbb{R}^2$), damping ($\bm{C} \in \mathbb{R}^2$), and static joint Coulomb friction ($\bm{\tau}_{ijf} \in \mathbb{R}^2$) defined at each joint:

\begin{equation}\label{eq:joint_torque}
  ^{ij}\bm{\tau}_{ij} =
  \begin{bmatrix}
    0 \\[4pt]
    -\bm{K}_{ij}\bm{\theta}_{ij}
    -\bm{C}_{ij}\dot{\bm{\theta}}_{ij}
    -{^{ij}\bm{\tau}_{ijf}}
  \end{bmatrix},
\end{equation}

where $\bm{K} = 0\bm{I}_{2\times2}$ due to assumption \ref{ass:inelastic}. Here, the components of $\bm{\theta}_{ij}$ describe the angle of link $j+1$ relative to link $j$ around $j$'s $Y$ and $Z$ axes respectively. It is straightforward to show that if the same stiffness $k = k_Y = k_Z$ and damping factors $c = c_Y = c_Z$ are selected around each of these axes (i.e. $\bm{K} = k\bm{I}_{2\times2}$ and $\bm{C} = c\bm{I}_{2\times2}$), the joint will display isotropic stiffness and damping properties. The same is not true for static joint friction $\bm{\tau}_{ijf}$, which is relevant in Section~\ref{sect:results}.

\emph{Remark 1}. If the massless rigid-rod assumption were adopted (as in \cite{sreenath_dynamics_2013, fink_planning_2011, michael_cooperative_2011, lee_geometric_2013, wu_geometric_2014}), each cable would reduce to a single element with $n_i = 1 \;\forall\,i$ and the internal forces would satisfy $\bm{f}_{i0} = -\bm{f}_{i(n_i+1)}$. While convenient, this assumption breaks down when inertial effects of the cable are significant, for example during aggressive maneuvers, when tension is re-engaged, or when the cable mass is non-negligible relative to the payload or the vehicles.

\emph{Remark 2}. By concentrating cable mass at the joints, prior works \cite{kotaru_differential-flatness_2018, kotaru_multiple_2020, goodarzi_stabilization_2015} avoid modeling link inertias, since all forces act through each link's concentrated point mass. This choice reduces the state representation: instead of tracking full link orientations $\bm{R}_{ij}$, the model only needs to evolve link direction vectors $\bm{q}_{ij} \in S^2 := \{ \bm{q} \in \mathbb{R}^3 \mid \bm{q}^\top \bm{q} = 1\}$. This effectively treats the cable as a chain of point masses, thus eliminating (\ref{eq:eom_cable_ang}), and not requiring the joints to provide balancing moments: $^{ij}\bm{\tau}_{ij} = 0$. The force resolution also simplifies, because forces transmitted through the rods are restricted to pure axial tension, $t_{ij} = f_{ij} \in \mathbb{R}$, which must lie along the known directions $\bm{q}_{ij}$. These simplifications are not available in our formulation, as the ABA requires rigid bodies with inertia. However, the ABA achieves efficiency by design, solving the system in $\mathcal{O}(n)$ time without further simplifications.

\emph{Remark 3.} Careful construction of the cable model's link directionality means that kinematic equations (\ref{eq:cable_load_end}) - (\ref{eq:cable_drone_end}) describe a tree-structured open kinematic chain with a single root at the load. This allows the ABA to solve link poses without explicitly computing internal forces $\bm{f}_{ij}$.

\emph{Remark 4.} As previously stated, it follows from \ref{ass:twisting_moment} that the moment around each cable joint's $X$ axis is \SI{0}{\newton\metre}. This allows these joints to be modeled as universal joints rather than ball joints, which decreases each joint's number of degrees of freedom, and thus complexity, by $1$. 

\emph{Remark 5.} This model does not assume that the cables are attached at each drone's center of mass in contrast to prior works \cite{sreenath_dynamics_2013, kotaru_differential-flatness_2018, li_rotortm_2024}. This provides flexibility to attach the cable where practical in real-world experiments, at the expense of the additional term ${^i\bm{r}_{i}} \times \bm{R}^{\intercal}_i\bm{f}_{in_i}$ in (\ref{eq:eom_drones_ang}).

\section{EXPERIMENTAL PLATFORM}\label{sect:method}
\begin{figure}[htbp]
    \centerline{\includegraphics[width=1\columnwidth]{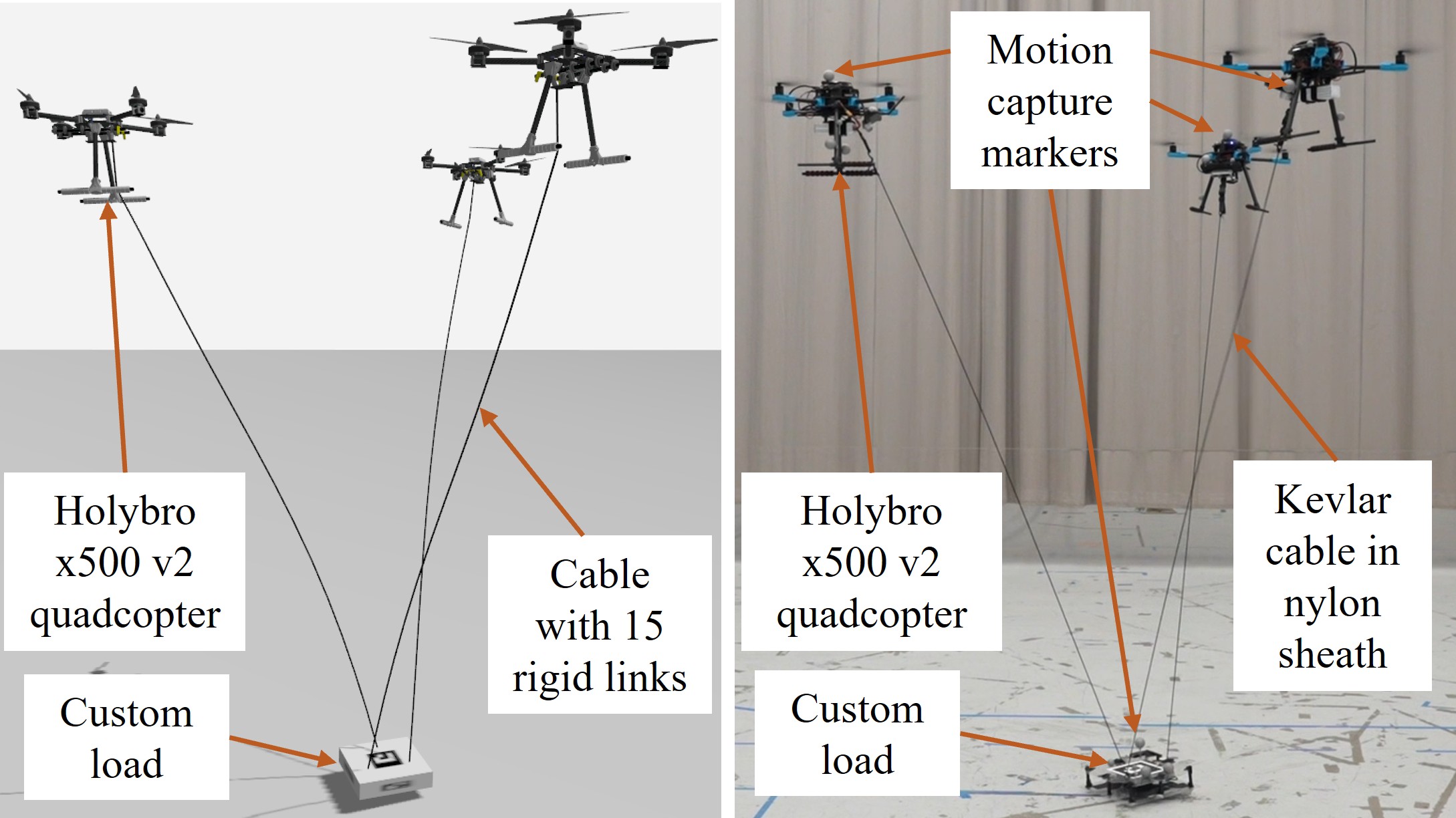}}
    \caption{Key elements of a three-quadrotor MUSLS in simulation (left) and the real world (right).} 
    \label{fig:system_method}
\end{figure}

In the previous section, we presented the dynamical equations of a MUSLS with a multi-link cable model. We now describe the experimental setup used to evaluate this model, implemented with the robotic system illustrated in Fig.~\ref{fig:system_method}.


The system is designed to transport the load to a desired pose, denoted $^W\bm{T}_{Ld}$, which can vary over time to define a load trajectory. To achieve this, we employ a \emph{position-based control scheme} where each drone independently tracks its own waypoint without considering inter-agent coupling or formation feedback \cite{oh_survey_2015}. This deliberately simple scheme is chosen because the focus of this work is on evaluating the cable dynamics model, not the controller; omitting load-pose feedback and inter-agent coupling exposes the intrinsic load and cable dynamics rather than obscuring them with a higher-level tracking controller as in \cite{li_rotortm_2024}.

To construct the drones' waypoints, a virtual inverted cone is used to specify $^{Ld}\bm{T}_{id}$, the desired pose of drone $i$ relative to the load's desired pose, such that the overall formation realizes $^W\bm{T}_{Ld}$. The cone’s tip is located at the load’s desired center of mass, while its base forms a circle (dotted line in Fig.~\ref{fig:system_symbols}) along whose circumference the drones are evenly spaced. Combining $^{Ld}\bm{T}_{id}$ with $^W\bm{T}_{Ld}$ yields the world-frame target pose (waypoint) for drone $i$:
\begin{equation}\label{eq:drone_pose_multiagent_coordination}
    {^W\bm{T}_{id}} = {^W\bm{T}_{Ld}} {^{Ld}\bm{T}_{id}}.
\end{equation}

\begin{figure}[htbp]
    \centerline{\includegraphics[width=1\columnwidth]{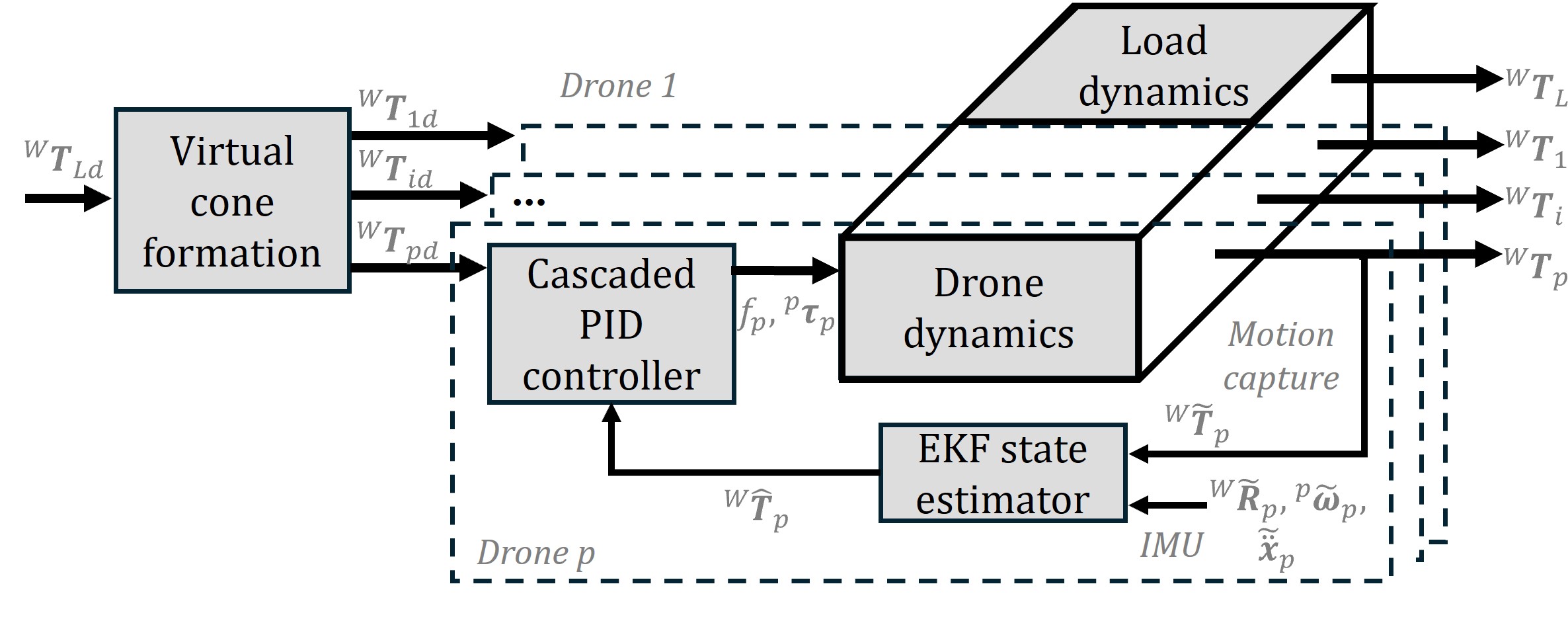}}
    \caption{Block diagram of the full MUSLS used for testing the dynamics model. Each drone has an onboard lower-level pose controller and state estimator that closes the loop with an individual drone's dynamics. The drones dynamics are coupled by the load and cables to form the full MUSLS. Measured quantities are indicated with $\overset{\sim}{.}$, while estimated quantities are indicated with $\hat{\cdot}$.}
    \label{fig:system_block_diagram}
\end{figure}

Fig.~\ref{fig:system_block_diagram} shows how each drone $i$ (corresponding to drone $i$ in Fig.~\ref{fig:system_symbols}) takes in the waypoint $^W\bm{T}_{id}$ generated using the virtual cone. Each waypoint $^W\bm{T}_{id}$ sets the setpoint of drone $i$'s onboard cascaded PID pose controller which regulates, from inner to outer loop, the angular velocity $\bm{\omega}_i$, attitude $\bm{R}_i$, linear velocity $\dot{\bm{x}}_i$, and position $\bm{x}_i$ of drone $i$ through commanded thrust $f_i$ and body torque $^i\bm{\tau}_i$. Because the drones are coupled through the load and cables, these inputs influence the entire MUSLS. Ground-truth measurements (denoted $\overset{\sim}{.}$) of the drones $^W\overset{\sim}{\bm{T}}_i$ and load $^W\overset{\sim}{\bm{T}}_L$ are obtained either from an external motion-capture system in hardware experiments or directly from simulator feedback. Each drone fuses these measurements with inertial measurement unit (IMU) data using an onboard extended Kalman filter (EKF). Low-level flight control and state estimation are provided by the open-source PX4 autopilot \cite{meier_px4_2015}, both in real-world experiments and in simulation. In the latter case, PX4 software-in-the-loop (SITL) is connected to Gazebo to fly the vehicles virtually. The controller gains used in the real world and in simulation are identical.

The parameters used for both simulated and real-world MUSLS seen in Fig. \ref{fig:system_method} can be found in Table \ref{tab:musls_params}. All real-world experiments are run on Holybro x500 v2 drone platforms equipped with \SI{5500}{\milli\ampere\hour} LiPo batteries, Pixhawk 6X flight controllers and Intel NUC7i5BNKs running Ubuntu 22.04. The NUCs are connected through ROS2 running over a TP-Link Archer AX11000 router, over which they receive motion capture data at \SI{100}{\hertz} from 20 Vicon motion capture cameras. All simulations are run on an AMD Ryzen 7 3700x using the DART dynamics engine in Gazebo \cite{lee_dart_2018}.



\begin{table}[htbp]
\centering
\caption{MUSLS parameters.}
\resizebox{\columnwidth}{!}{%
\begin{tabular}{l l}
\toprule
\textbf{Load} & \\
Mass $m_L$ (\unit{\kilo\gram}) & 1.66 \\
Inertia $\bm{J}_L$ (\unit{\kilo\gram\metre\squared}) &
$\begin{bmatrix}
0.0101956 & 0.0003278 & -0.000108.0 \\
0.0003278   & 0.013750 & -0.0003215 \\
-0.0001080  & -0.0003215  & 0.0209970
\end{bmatrix}$ \\
Cable 1 attach. pt. $^L\bm{r}_{1L}$ (\unit{\metre}) & $[0.1, 0, 0.03]^{\intercal}$ \\
Cable 2 attach. pt. $^L\bm{r}_{2L}$ (\unit{\metre}) & $[-0.05, 0.087, 0.03]^{\intercal}$ \\
Cable 3 attach. pt. $^L\bm{r}_{3L}$ (\unit{\metre}) & $[0.03]^{\intercal}$ \\
\midrule
\textbf{Drones} & \\
Mass $m_i$ (\unit{\kilo\gram}) & 2.11 \\
Inertia $\bm{J}_i$ (\unit{\kilo\gram\metre\squared}) &
$\begin{bmatrix}
0.0216667 & 0.0 & 0.0 \\
0.0 & 0.0216667 & 0.0 \\
0.0 & 0.0 & 0.040000
\end{bmatrix}$ \\
Cable attach. pt. $^i\bm{r}_{i}$ (\unit{\metre}) & $[0.04, 0, 0]^{\intercal}$ \\
\midrule
\textbf{Cables} & \\
Mass $m_{ic}$ (\unit{\kilo\gram}) & 0.02 \\
Length $l_{ic}$ (\unit{\metre}) & 2.42 \\
Radius $r_{ic}$ (\unit{\metre}) & 0.0031 \\
\bottomrule
\end{tabular}
} 
\label{tab:musls_params}
\end{table}

\section{RESULTS}\label{sect:results}
\begin{figure*}[htbp]
    \centerline{\includegraphics[width=1\textwidth, height=0.25\textheight]{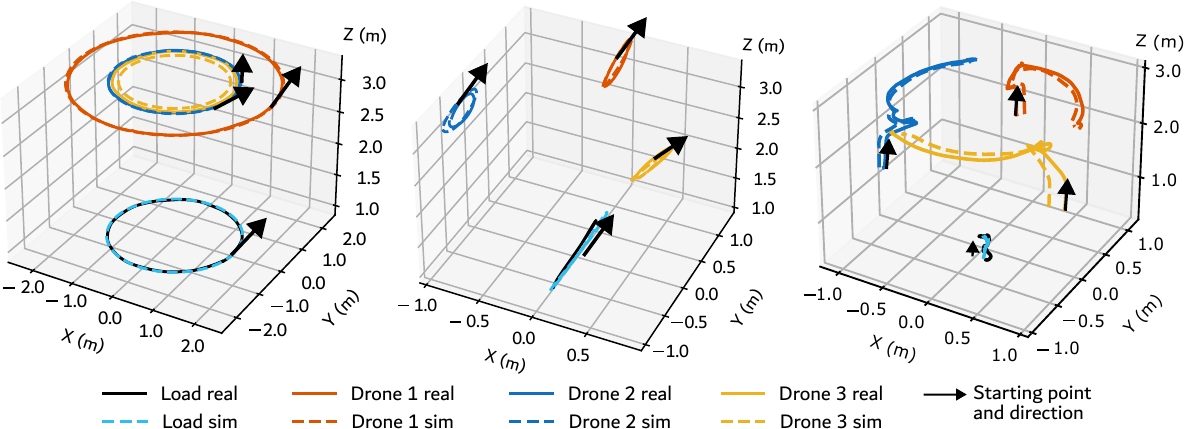}}
    \caption{Comparison of mean drone and load trajectories in a MUSLS. `Real' trajectories are generated experimentally in the real world, while `sim' trajectories are generated in simulation using the tuned multi-rigid-link cable model. The three sets of trajectories shown are (from left to right): a quasi-static circle, one period of simple harmonic motion along the $Y$-axis, and one cycle of cable tension engagement.} 
    \label{fig:results_3d}
\end{figure*}

To quantify the accuracy of the dynamics model presented in Section~\ref{sect:dynamics}, we command the simulated and real-world MUSLSs described in Section~\ref{sect:method} to follow three desired load trajectories (shown in Fig.~\ref{fig:results_3d}):

\begin{itemize}
    \item a circle of radius \SI{1.5}{\metre} at a tangential velocity of \SI{0.3}{\metre\per\second} with the load's x-axis always pointing radially out from the center of the circle (a simple quasi-static trajectory),
    \item simple harmonic motion (SHM) of amplitude \SI{1}{\metre} and angular frequency of \SI{2}{\radian\per\second} along the inertial frame's y-axis (a more aggressive trajectory), and
    \item a rapid increase in height of \SI{0.5}{\metre} and yaw of \SI{1.7}{\radian} after previously resting on the ground (testing cable tension engagement modeling).
\end{itemize}
A detailed discussion of the performance of the final tuned model is given in Section~\ref{sect:results_final}, following the parameter sensitivity study in Section~\ref{sect:param_sense}. The code and data needed to reproduce these results, including rosbags from the real-world MUSLS flights, are provided in the accompanying repository~\cite{harvey_merton_code_2025}.


\emph{Error metrics:} Agreement between real-world and simulated trajectories is quantified using translational and rotational error measures. The translational error is given by
\begin{equation}\label{eq:pos_err}
    \|{^W\bm{x}_{\text{dist}}}\| = \|{^W\bm{x}_{b}} - {^W\bm{x}_{a}}\|,
\end{equation}
and the rotational error is defined as the geodesic distance
\begin{equation}\label{eq:orient_err}
    \text{dist}_{\theta}({^W\bm{R}_{b}},{^W\bm{R}_{a}}) 
    = \cos^{-1}\!\left(\tfrac{\text{tr}(({^W\bm{R}_{b}})^{\intercal}{^W\bm{R}_{a}})-1}{2}\right).
\end{equation}
The mean error over a trajectory is obtained by first computing the error at each timestep and then averaging across the entire trajectory. Although this provides a useful performance metric, it can overstate discrepancies in oscillatory motions (e.g., SHM translation), since minor period mismatches produce large instantaneous amplitude differences.


\emph{Results processing:} Each full flight (both in simulation and in the real world) comprises two full circles, seven full SHM periods, and three cycles of tension engagement. In the real world and for final simulations, the full flight is repeated five times, giving a total of ten circles, thirty-five SHM periods, and fifteen cycles of tension engagement. Of the real-world data, two circles and one yaw engage were rejected due to anomalously large communication dropout affecting the desired drone arrangement. The results from simulation in Section~\ref{sect:param_sense} come from one full flight each.

Each logfile is sliced according to the start and end of each periodic motion. The slices from all five runs are then averaged to produce the mean trajectory. \SI{95}{\percent} confidence intervals are constructed by invoking the central limit theorem on the distribution of the mean trajectory.

\subsection{Parameter sensitivity study}\label{sect:param_sense}
As shown by (\ref{eq:joint_torque}), it is necessary to specify the damping and friction of each joint, along with the number of cable elements. The selection procedure is as follows:

\begin{enumerate}
    \item Set the number of cable elements to $29$, which is the maximum supported by the simulator. For larger values, DART fails to resolve the system kinematics.
    \item Specify the damping coefficient of the cable joints.
    \item Specify the static friction value of the cable joints.
    \item Conduct a sensitivity analysis on the number of cable elements to determine the final element count.
\end{enumerate}

\begin{figure}[htbp]
    \centering
    \subfloat[Joint damping factor sensitivity with \SI{0}{\newton\metre} of joint friction.\label{fig:results_sense_damping}]{
        \includegraphics[width=1\columnwidth]{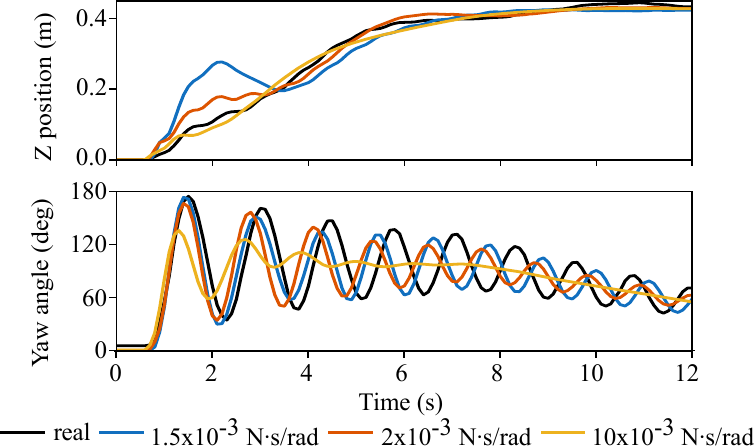}
    }\\[0.5em]  
    \subfloat[Joint friction sensitivity with \SI{2e-3}{\newton\second\per\radian} of joint damping.\label{fig:results_sense_friction}]{
        \includegraphics[width=1\columnwidth]{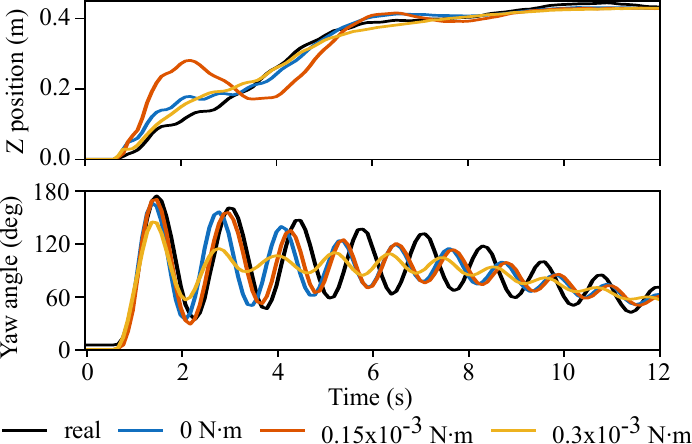}
    }
    \caption{Z position and yaw angle of the load during the tension engage trajectory while varying the simulated: (a) joint damping, (b) joint friction. The real-world load's trajectory is provided in black for reference. All experiments are performed with 29 cable elements and \SI{0}{\newton\metre\per\radian} of joint stiffness.}
    \label{fig:results_sense_damping_frict}
\end{figure}

\emph{Joint damping:} Fig.~\ref{fig:results_sense_damping} shows that reducing the damping factor improves the match between simulated and real-world load yaw oscillations during the tension-engagement trajectory. However, excessively low damping values lead to unstable simulations. This is evident from the peak observed in the load’s $Z$ position during tension engagement at approximately \SI{2}{\second} with a damping factor of \SI{1.5e-3}{\newton\second\per\radian}. To balance stability with fidelity, a damping factor of \SI{2e-3}{\newton\second\per\radian} is selected.

\emph{Joint friction:} Fig. \ref{fig:results_sense_friction} demonstrates that increasing the static friction of the joints can reduce the stability of the tension engagement simulation (with a $Z$-position peak at \SI{\sim 2}{\second} for \SI{0.15e-3}{\newton\metre}) and can also introduce irregular oscillatory behavior in the yaw response (notably at \SI{0.3e-3}{\newton\metre}). This effect is likely exacerbated by the anisotropic joint friction described in Section~\ref{sect:dynamics}. Moreover, including joint friction substantially increases computational demand, reducing simulation speed from approximately \SI{99}{\percent} to \SIrange[range-phrase = --]{40}{56}{\percent} of real time (depending on the number of cable elements, as seen in Table~\ref{tab:runtime}). To preserve both stability and efficiency, the static joint friction is therefore set to \SI{0}{\newton\metre}.

\begin{table}[h]
\centering
\caption{Run speed of the MUSLS simulator under different cable settings. Maximum step size \SI{4e-3}{\second}.\textsuperscript{*}}
\begin{tabular}{lc}
\toprule
\textbf{Setting} & \textbf{Realtime factor (approx.)} \\
\midrule
15 links, no friction   & \SI{99}{\percent} \\
15 links, with friction & \SIrange[range-phrase = --]{94}{98}{\percent}  \\
29 links, no friction   & \SI{99}{\percent} \\
29 links, with friction & \SIrange[range-phrase = --]{40}{56}{\percent} \\
\bottomrule
\multicolumn{2}{l}{\footnotesize\textsuperscript{*}Tension engagement excluded: requires a timestep of} \\
\multicolumn{2}{l}{\footnotesize \SI{2e-4}{\second} to maintain stability.} 
\end{tabular}
\label{tab:runtime}
\end{table}

\begin{figure}[htpb]
    \centering
    \includegraphics[width=0.85\columnwidth]{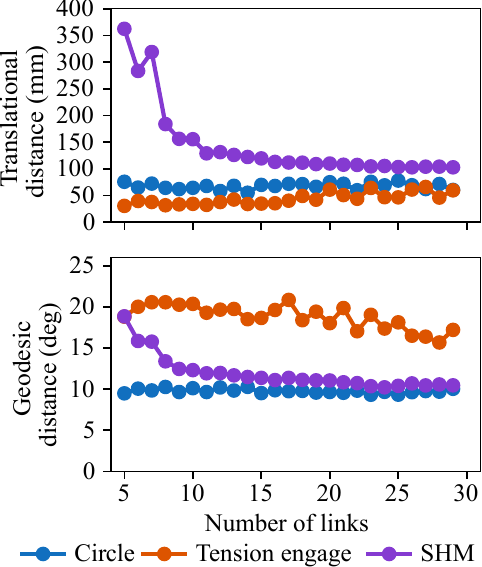}
    \caption{Mean distance (over each trajectory) of the mean simulated load's pose to the mean real-world load's pose under a varied number of cable elements. Joint damping: \SI{2e-3}{\newton\second\per\radian}, joint friction: \SI{0}{\newton\metre}.}
    \label{fig:results_sense_num_elems}
\end{figure}

\emph{Number of elements:} Fig.~\ref{fig:results_sense_num_elems} shows that increasing the number of cable elements reduces the mean load pose error for the SHM trajectory up to about 25 elements. For the tension-engagement trajectory, however, using more than 15 elements leads to highly variable angular accuracy and generally increases load position error, reflecting the instability introduced at higher element counts. Conversely, using fewer than five elements causes instabilities in the SHM trajectory that result in a crash. To ensure both stability and accuracy, each cable is therefore modeled with 15 elements.


Note in Fig.~\ref{fig:results_sense_num_elems} that the mean SHM translational error and tension-engagement geodesic error are larger than those of the other trajectories. This reflects the oscillatory nature of these trajectories, which, as discussed when presenting the error metrics, tends to amplify average error values.


\subsection{Performance evaluation}\label{sect:results_final}
\begin{figure*}[htpb]
    \centering
    \includegraphics[width=1\textwidth, height=0.62\textheight]{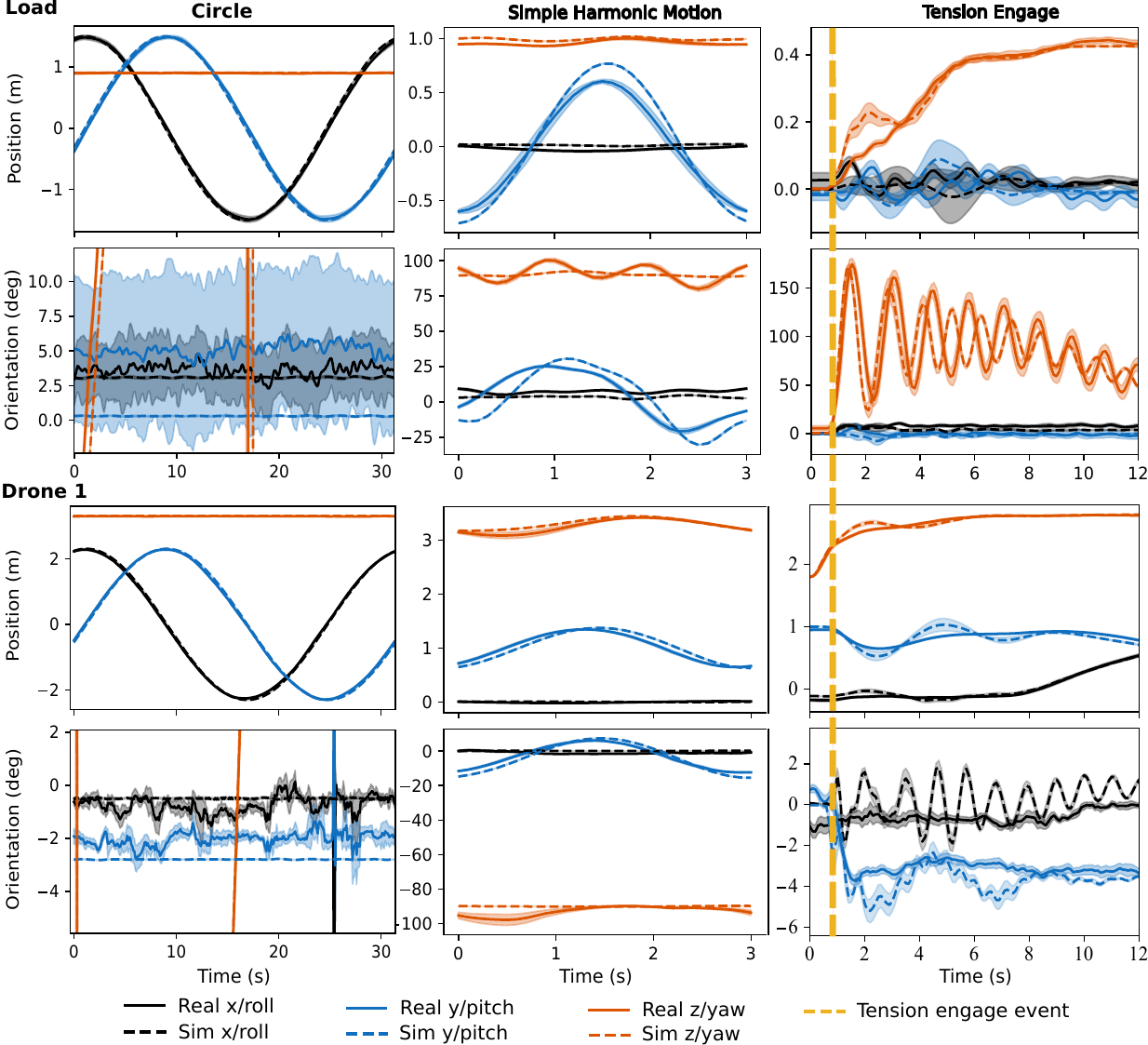}
    \caption{Mean position and orientation of the load and drone 1 in the final tuned MUSLS model across the trajectories shown in Fig. \ref{fig:results_3d}. For the circular trajectory, yaw is clipped for both the load and drone 1 to highlight roll and pitch. The shaded regions represent \SI{95}{\percent} confidence intervals. }
    \label{fig:results_components}
\end{figure*}

From Sections \ref{sect:dynamics} and \ref{sect:param_sense}, the final selected cable parameters are: 

\begin{itemize}
    \item 15 elements each,
    \item joint stiffness of \SI{0}{\newton\metre\per\radian},
    \item joint damping factor of \SI{2e-3}{\newton\second\per\radian}, and
    \item joint friction of \SI{0}{\newton\metre},
\end{itemize}
where each joint is modeled with a universal joint. Following the trajectories shown in Fig.~\ref{fig:results_3d}, Fig.~\ref{fig:results_components} compares the pose of the load and drone 1 between the simulated and real-world MUSLS under the selected cable parameters. We see that both the simulated load and drone 1 mostly follow the corresponding real-world poses across all trajectories. 

Despite the overall agreement between simulation and the real world, Fig.~\ref{fig:results_components} highlights several notable discrepancies:

\begin{itemize}
    \item large real-world spikes in drone 1's roll and pitch around \SI{25}{\second} as drone 1 momentarily lost communication and dropped out of formation during one circle flight,
    \item a larger simulated load SHM amplitude in $Y$ than is achieved in the real world,
    \item a significantly modified simulated load pitch during the SHM trajectory,
    \item a shorter simulated load yaw oscillation period with faster damping during tension engage,
    \item higher simulated drone 1 oscillations in roll and pitch during tension engage, and
    \item a $Z$ peak for the load during tension engage around \SI{2}{\second}.
\end{itemize}
The second through fifth discrepancies are likely attributable to imperfect modeling of the load or cable inertial properties, since the simulated drone trajectory aligns well with the real-world SHM motion, though further testing is needed to confirm this hypothesis. On the other hand, the final discrepancy reflects the multi-link cable model's inability to successfully capture cable tension engagement, reinforcing the need for enhanced engagement modeling such as hybrid dynamics approaches \cite{li_rotortm_2024}.

The wider confidence intervals in the real-world trajectories are likely due to higher communication latency, which introduces small inconsistencies in the drones’ commanded pose updates and, in turn, increases MUSLS pose variability across flights. An exception occurs for the load and drones' positions during the tension engagement phase, where the broad simulated intervals highlight the model's inherent variability. This variability arises because small changes in link orientation can significantly alter the impulse transmitted to the load during cable tension engagement, again reinforcing the need for more advanced tension engagement models.

Table~\ref{tab:results_mean_error} reports the translational and geodesic errors between mean simulated and real-world trajectories. The largest geodesic error, \SI{17.2}{\degree}, occurs during the tension-engagement trajectory, compared with a maximum of \SI{11.4}{\degree} in other cases. This discrepancy reflects a period mismatch between simulated and real-world yaw oscillations, which, as discussed earlier, artificially increases average errors. The period mismatch likely stems from inaccuracies in the modeled load properties and may be mitigated in future experiments through more precise inertia characterization. 

\begin{table}[htpb]
\centering
\caption{Mean translational (Trans.) and geodesic (Geo.) error of the simulated MUSLS.}
\begin{tabular}{lcccc}
\toprule
& \multicolumn{2}{c}{\textbf{Circle}} & \multicolumn{2}{c}{\textbf{SHM}} \\
\cmidrule(lr){2-3} \cmidrule(lr){4-5}
Agent & Trans. (mm) & Geo. (deg) & Trans. (mm) & Geo. (deg) \\
\midrule
Load    & 67.8   & 9.7   & \textbf{118.5}  & 11.4  \\
Drone 1 & 74.8   & 1.7   & 95.1   & \textbf{4.1}   \\
Drone 2 & 68.3   & 1.8   & 118.0  & 3.0   \\
Drone 3 & \textbf{204.1}  & 2.1   & 110.5  & 2.6   \\
\midrule
& & \multicolumn{2}{c}{\textbf{Engage}} \\
\cmidrule(lr){3-4}
Agent   & & Trans. (mm) & Geo. (deg) & \\
\midrule
Load    & & 38.7   & \textbf{17.2}  & \\
Drone 1 & & 78.0   & 1.3   & \\
Drone 2 & & 43.5   & 1.6   & \\
Drone 3 & & 132.4  & 1.4   & \\
\bottomrule
\end{tabular}
\label{tab:results_mean_error}
\end{table}

For translation, all load and most drone trajectories maintain errors below \SI{132.4}{\milli\metre} (approximately \SI{5.5}{\percent} of the cable length). The only exception is drone 3 in the circular trajectory, which exhibits an error of \SI{8.4}{\percent} of the cable length. This is due to drone 3 following a trajectory of slightly smaller radius in simulation than the real world (see Fig. \ref{fig:results_3d}). Again, this is likely due to model parameter inaccuracy, but further testing is required to confirm this hypothesis. Recall that our setup does not use load pose feedback, so results are not directly comparable to prior closed-loop approaches.

\section{CONCLUSION}\label{sect:conclusions}
In this work, we presented a dynamical model of a multi-UAV slung-load system using a multi-rigid-link cable formulation that is efficiently solvable via Featherstone’s algorithm. Through a systematic sensitivity analysis, we identified a simple set of parameters that balance fidelity and efficiency: joint stiffness of \SI{0}{\newton\metre\per\radian}, damping of \SI{2e-3}{\newton\second\per\radian}, joint friction of \SI{0}{\newton\metre}, and a discretization of 15 links per cable. With this configuration, the model reproduces real-world behavior at real-time speed, achieving orientation errors below \SI{11.4}{\degree} and translational errors of approximately \SI{5.5}{\percent} of the cable length.

Future work will focus on directly comparing this model to the massless rigid rod assumption, and determining where that assumption is valid. Work will also focus on capturing more advanced cable dynamics, particularly to improve tension engagement prediction and performance under aggressive maneuvers. To facilitate such continued research, real-world MUSLS flight data has been made available.

\addtolength{\textheight}{-8cm}   








\bibliographystyle{IEEEtran}
\bibliography{Bibliography.bib}

\end{document}